\documentclass[letterpaper, 10 pt, conference]{ieeeconf}  

\IEEEoverridecommandlockouts                              

\usepackage{color}
\usepackage{graphicx,amsmath,amssymb,bm,xcolor}
\usepackage{subcaption} 
\usepackage{multirow} 
\usepackage[utf8]{inputenc}
\usepackage{breqn}
\usepackage[noadjust]{cite}

\usepackage{cleveref}

\newcommand{\trsp}{{\scriptscriptstyle\top}}

\DeclareMathOperator*{\argmin}{argmin} 

\title{\LARGE \bf
Active Improvement of Control Policies \\ with Bayesian Gaussian Mixture Model
}

\author{Hakan Girgin, Emmanuel Pignat, No\'emie Jaquier and Sylvain Calinon
\thanks{All the authors are with the Idiap Research Institute, Rue Marconi 19, 1920 Martigny, Switzerland (e-mail: \tt\scriptsize firstname.lastname@idiap.ch).}%
\thanks{*This work was supported by the CoLLaboratE project (https://collaborate-project.eu/), funded by the EU within H2020-DT-FOF-02-2018 under grant agreement 820767, and by the ROSALIS project (Swiss National Science Foundation).}
}

\begin{document}

\maketitle
\thispagestyle{empty}
\pagestyle{empty}

\begin{abstract}
Learning from demonstration (LfD) is an intuitive framework allowing non-expert users to easily (re-)program robots. However, the quality and quantity of demonstrations have a great influence on the generalization performances of LfD approaches.
In this paper, we introduce a novel active learning framework in order to improve the generalization capabilities of control policies. The proposed approach is based on the epistemic uncertainties of Bayesian Gaussian mixture models (BGMMs). We determine the new query point location by optimizing a closed-form information-density cost based on the quadratic Rényi entropy. Furthermore, to better represent uncertain regions and to avoid local optima problem, we propose to approximate the active learning cost with a Gaussian mixture model (GMM).
We demonstrate our active learning framework in the context of a reaching task in a cluttered environment with an illustrative toy example and a real experiment with a Panda robot.
\end{abstract}

\section{INTRODUCTION}

Learning from demonstration (LfD) offers an intuitive framework to overcome the difficulty of programming robots by teaching them movements using an adaptive representation. One of the main LfD approaches is called behavior cloning or policy imitation, and consists in inferring the parameters of a movement model via supervised learning \cite{atkeson1997locally,ijspeert02NIPS,Paraschos13,calinon2007learning,cederborg2010incremental,Khansari11TRO,levine2016end} from a demonstration dataset. In LfD, the demonstrations are often acquired by kinesthetic teaching or by teleoperation. One of the main advantages of these techniques is that they allow non-expert users to easily (re-)program the robots. However, it may not be straightforward to determine the number of demonstrations necessary for the robot to learn a specific skill, as well as the locations in which they should be provided. Moreover, acquiring the demonstrations can be costly especially in industrial environments. Therefore, we aim at collecting these demonstrations in an informative way.

Active learning is a promising approach to address the aforementioned issues. An active learning framework develops and tests new hypotheses in an interactive learning process. In robotics, the robot is first provided with initial demonstrations from which an initial model of the task can be built. Then, at each stage of the active learning framework, the robot is expected to request a new demonstration in order to improve the model. This contrasts with passive learning systems that attempt to explain the model only according to available training data. Ideally, the robot should request the new demonstration around a query point that will maximize the information gain. Specifically, the information gain is related to the areas where the model uncertainties are the highest. 

\begin{figure}[tbp]
	\centering
	\includegraphics[width=.8\columnwidth]{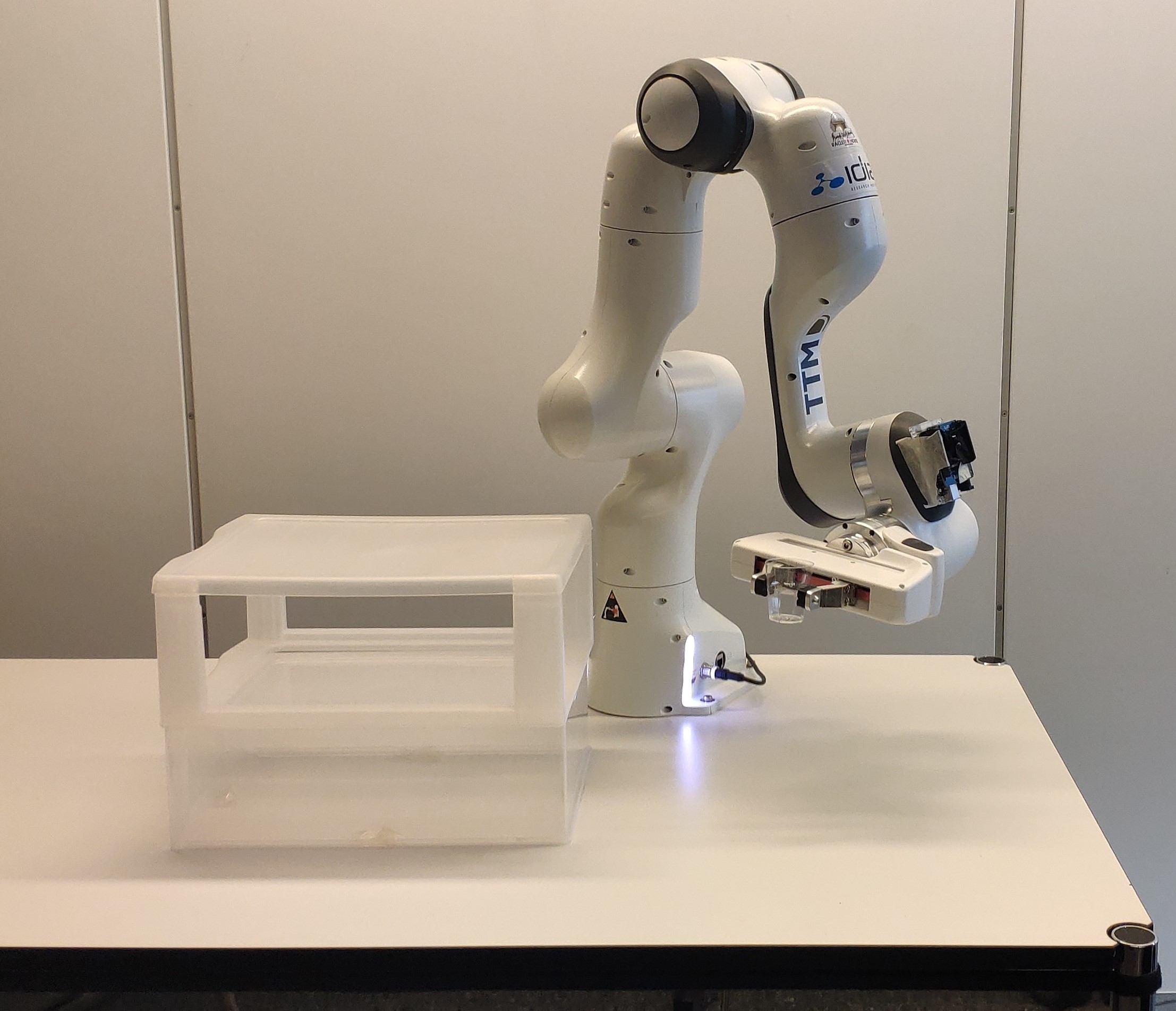}
	\caption{Experimental setup with Franka Emika Panda robot. The task is to put the cup inside a box covered from top and bottom, starting from anywhere in the space. The robot has to maintain a specific end-effector orientation to perform the task without pouring the cup. The main challenge is not to collide with the box and the other obstacles in the environment.}
	\label{fig:setup}
\end{figure}
In robotics applications, two different kinds of uncertainties arise, namely \emph{(i)} \textit{aleatoric uncertainties} and \emph{(ii)} \textit{epistemic uncertainties}. The aleatoric uncertainties represent the variations in the demonstrations and are typically used to adapt the behavior of the robot, e.g. its compliance at different phases of the task. In contrast, the epistemic uncertainties are related to the lack of knowledge (i.e. data) in the demonstrations and is typically used for informative exploration. Active learning is thus based on the epistemic uncertainties in the model. A natural way to take these uncertainties into account is through Bayesian inference \cite{bishop2006}.

In this work, we propose an active learning approach with the aim of improving the generalization capabilities of control policies in a behavior cloning setup, also called policy imitation. Our approach is based on the framework presented in \cite{Pignat2019BayesianGM} which models a joint distribution $p(\bm{x}_t, \bm{u}_t)$ in action-state abstraction with Bayesian Gaussian mixture models (BGMMs). The conditional (predictive) distribution of the policy $p(\bm{u}_t|\bm{x}_t)$ is then found by conditioning on the current state $\bm{x}_t$. In \cite{Pignat2019BayesianGM}, the authors use a product of experts (PoE) framework to exploit the uncertainties inherent to Bayesian models in order to fuse several control policies (see Section~\ref{sec:Bckgr} for a brief background). The proposed active learning approach is based on the epistemic uncertainties in BGMM model. A method to decompose the covariance matrix of the posterior BGMM distribution into aleatoric and epistemic parts is first presented in Section~\ref{subsec:Uncertainty}. The quadratic Rényi entropy is then used to compute the related uncertainties of Gaussian mixture models (GMMs) in closed-form (see Section~\ref{subsec:Renyi}). As explained in Section~\ref{subsec:Cost}, the next query point of our active learning framework is obtained by maximizing an information-density cost based on the quadratic Rényi entropies. In particular, we propose to approximate this cost with a GMM to represent highly uncertain region distribution. This notably avoids local optima problem during uncertainty maximization. Finally, we demonstrate the efficiency of our approach on a reaching task in a cluttered environment in a 2D simulated example and with a real experiment on a Panda robot (see Section~\ref{sec:Experiments}). The experiment setup is presented by Fig.~\ref{fig:setup}.

The contributions of this paper are threefold: \emph{(i)} we provide an uncertainty decomposition in BGMM control policies to be used for exploitation and exploration in behavior cloning approaches; \emph{(ii)} we introduce an information weighted closed-form cost to describe uncertain regions of the state space; and \emph{(iii)} we propose an active learning framework which can be used with partial demonstrations, with closed-form monitoring of the uncertainty reduction.

\section{RELATED WORK}
\label{sec:RelatedWork}
A collection of recent work focuses on improving and fine-tuning learned movement representations using reinforcement learning (RL) \cite{kober2013survey, deisenroth2013survey} and iterative learning control (ILC) \cite{bristow2006survey}. As these methods iteratively minimize a reward function, LfD can be used to determine the initial point of the optimization in order to favor a safe exploration.
In contrast, information-theoretic explorations in behavior cloning methods have been exploited only in few works to enhance the quality and the generalization abilities of the learned movement models \cite{Maeda17,kroemer2010, Conkey2019}.

One of the simplest and widely used active learning methods is uncertainty sampling. Using an uncertainty measure, the robot is expected to request a query point in the most uncertain region of the input space. If the model can only encode aleatoric uncertainties, one can train several probabilistic representations with different local convergence properties. The disagreement between each individual model and their average model is then maximized using KL divergence as explained in \cite{settles2012active}. Other techniques consist in reducing the variance of error in a regression problem. In general, this is intractable. Simplifications occur by using Fisher Information and Cramér-Rao inequality as in \cite{abraham2019active}. All the aforementioned methods are myopic as they only care about the information content of single data instances. This can result in models selecting outliers or exploring far away in the context space where no generalization is required. Information-density methods overcome this problem by choosing instances that have high information content and are still representative of the underlying distribution. This is achieved by using a weighted product of uncertainty measure (entropy, ensemble, etc) and similarity  measure (Euclidean distance, correlation coefficients, etc.) \cite{settles2012active}.

In \cite{Maeda17}, the authors use Gaussian Process Regression (GPR) in a reaching task to map object positions to the weights encoding the trajectories via Dynamic Movement Primitives (DMP). They demonstrate an active learning framework based on the GPR epistemic uncertainties for reaching to a predefined set of object positions to improve their DMP model. They work with time-dependent trajectory policies without control information. They measure the epistemic uncertainty of a whole trajectory given a context, while  aleatoric uncertainties (variations) are not considered. Our work differs from \cite{Maeda17} in two ways. First, as we consider state-dependent policies including both aleatoric and epistemic uncertainties. Second, their approach in \cite{Maeda17} exploits uncertainty sampling, which would diverge if the uncertainty is defined over a continuous variable instead of a discrete set of variables. To overcome this problem, we use information-density methods.

\section{BACKGROUND}
\label{sec:Bckgr}
In this section, we present the BGMM framework exploited to learn control policies presented in \cite{Pignat2019BayesianGM}. As state-dependent control policies learned with BGMM can create unstable behaviors, the BGMM policy is fused with another stable control policy within the PoE framework.

\subsection{Bayesian Gaussian Mixture Model}
\label{subsec:BGMM}

\newcommand{\xy}{\bm{x}}
\newcommand{\xyt}{\bm{\hat{x}}}
\newcommand{\XY}{\bm{X}}
\newcommand{\inp}{i}
\newcommand{\out}{o}
\newcommand{\cond}{\out|\inp}
\newcommand{\xinp}{\bm{x}^\inp}
\newcommand{\xout}{\bm{x}^\out}

\newcommand{\xinpt}{{\xyt^{\inp}}}
\newcommand{\xoutt}{{\xyt^{\out}}}

\newcommand{\xinpttr}{{{\xyt^\inp}^\trsp}}
\newcommand{\xouttr}{{{\xyt^\out}^\trsp}}

\newcommand{\sumgmm}{\sum_{k=1}^{K}}
\newcommand{\sumgmmj}{\sum_{j=1}^{K}}
\newcommand{\prodgmm}{\prod_{k=1}^{K}}
\newcommand{\prodgmmj}{\prod_{j=1}^{K}}
\newcommand{\sumdata}{\sum_{n=1}^{N}}
\newcommand{\proddata}{\prod_{n=1}^{N}}

\newcommand{\Linv}{{\bm{L}_k^{\inp\inp}}^{-1}}
\newcommand{\gmr}{\bm{L}_k^{\out\out}-\bm{L}_k^{\out\inp}\Linv\bm{L}_k^{{\out\inp^\trsp}}}
In this section, the Bayesian analysis of a Gaussian Mixture Model (GMM) is treated following \cite{bishop2006}.
Let $\xy = \begin{bmatrix}{\xinp}^\trsp\, {\xout}^\trsp \end{bmatrix}^\trsp \in \mathbb{R}^{D} $ be the joint observation of the input and the output with dimension $D = D_\inp{+}D_\out$. The joint distribution is defined with a mixture of $K$ multivariate normal distributions (MVNs) with means $\bm{\mu}{=}\{\bm{\mu}_k\}$, precision matrices $\bm{\Lambda}{=}\{\bm{\Lambda}_k\}$ and mixing coefficients $\bm{\pi}{=}\{\pi_k\}$ as
\begin{equation*}
	p(\xy|\bm{\pi},\bm{\mu}, \bm{\Lambda}) = \sumgmm\pi_k\mathcal{N}(\xy|\bm{\mu}_k, \bm{\Lambda}_k^{-1}).
\end{equation*}
We define a latent variable $\bm{z}$, each component of which is a binary variable $z_k \in \{0,1\}$ such that $\sumgmm z_k=1$. We can associate the mixing coefficients to the latent variables with $p(z_k{=}1) = \pi_k$ so that $p(\bm{z}|\bm{\pi}) = \prodgmm\pi_k^{z_k}$. We then obtain $p(\xy|\bm{z},\bm{\mu}, \bm{\Lambda}) = \prodgmm\mathcal{N}(\xy|\bm{\mu}_k, \bm{\Lambda}_k^{-1})^{z_k}$. The conditional distributions $p(\bm{Z}|\bm{\pi})$, $p(\XY|\bm{Z}, \bm{\mu}, \bm{\Lambda})$, the conjugate prior distributions $p(\bm{\mu}, \bm{\Lambda})$ and $p(\bm{\pi})$ of the joint observation dataset $\XY{=}\{\xy_n\}$ and the latent variable dataset $\bm{Z}{=}\{\bm{z}_n\}$ are summarized in Table \ref{table:bgmm}.
\begin{table}[h!]
	\centering
	\caption{conditionals and priors where $\mathcal{W}$($\cdot$) and $\mathrm{Dir}$($\cdot$) correspond to Wishart and Dirichlet distributions }\label{table:bgmm}
	\resizebox{\linewidth}{!}{%
	\begin{tabular}{ c||c }
		\hline
		\textbf{Conditional of $\XY$} &  \multirow{2}{*}{ $\proddata\prodgmm\mathcal{N}(\xy_n|\bm{\mu}_k, \bm{\Lambda}_k^{-1})^{z_{nk}}$} \\
		$p(\XY|\bm{Z}, \bm{\mu}, \bm{\Lambda})$ & \\
		\hline
		\textbf{Conditional of $\bm{Z}$} &  \multirow{2}{*}{ $\proddata\prodgmm\pi_k^{z_{nk}}$} \\
		$p(\bm{Z}|\bm{\pi})$ & \\
		\hline
		\textbf{Prior on $\bm{\mu}, \bm{\Lambda}$ } & \multirow{2}{*}{$\prodgmm\mathcal{N}\big(\bm{\mu}_k|\bm{m}_0, (\beta_0 \bm{\Lambda}_k)^{-1}\big)\mathcal{W}(\bm{\Lambda}_k|\bm{W}_0,\nu_0)$} \\
		$p(\bm{\mu}, \bm{\Lambda})$ &   \\ 
		\hline
		\textbf{Prior on $\bm{\pi}$ } & \multirow{2}{*}{$\mathrm{Dir}(\bm{\pi}|\alpha_0)$} \\
		$p(\bm{\pi})$ &   \\ 
		\hline
	\end{tabular}%
	}
\end{table}

As explained in \cite{bishop2006}, closed-form update equations for Expectation-Maximization (EM) algorithm is derived by using a factorized variational distribution. Note that EM update equations are usually implemented in machine learning libraries such as \textit{scikit-learn} for Python.

For robotic applications, we determine the predictive density of a new observation point $\xyt=\begin{bmatrix}{\xyt^{\inp\trsp}}\, {\xyt^{\out\trsp}} \end{bmatrix}^\trsp$ equivalent to a mixture of multivariate t-distributions with mean $\bm{\hat{m}}_k$, covariance matrix $\bm{\hat{L}}_k$, mixing coefficients $\hat{\pi}_k$ and degree of freedoms $\hat{\nu}_k$ as \cite{bishop2006} 
\begin{align}
	p(\xyt|\bm{X})=\sumgmm \pi_k\text{t}(\xyt|\bm{m}_k, \bm{L}_k, \nu_k),
\end{align}
where
\begin{align}
	\pi_k&=\frac{\alpha_k}{\sumgmm \alpha_k},\\
	\nu_k&=\nu_k+1-D,\\
	\bm{L}_k&=\frac{(\nu_k+1-D)\beta_k}{1+\beta_k}\bm{W}_k,\\
	\bm{m}_k&=\bm{\bar{m}}_k.
\end{align}
with the update equations on $\alpha_k$,$\beta_k$ $\nu_k$, $\bm{W}_k$ and $\bm{\bar{m}}_k$ are given in \cite{bishop2006}. We can then define the distribution of the output conditioned on the input as 
\begin{equation}
p(\xoutt |\xinpt, \XY)=\sumgmm \pi_k^\cond \text{t}(\xinpt|\bm{m}_k^\cond, \bm{L}_k^\cond, \nu_k^\cond),
\end{equation}
where
\begin{align}
	\pi_k^\cond &=\frac{\pi_k\text{t}(\xinpt|\bm{m}_k^\inp, \bm{L}_k^\inp, \nu_k^\inp)}{\sumgmmj \pi_j\text{t}(\xinpt|\bm{m}_j^\inp, \bm{L}_j^\inp, \nu_j^\inp)},\\
	\nu_k^\cond&=\nu_k+D^\inp, \\
	\hat{m}_k^\cond&=\bm{m}_k^\out+\bm{L}_k^{\out\inp}\Linv(\xinpt-\bm{m}_k^\inp), \\
	\bm{L}_s &= \gmr, \\
	L_k^\cond&=\frac{\nu_k + (\xinpt-\bm{m}_k^\inp)^\trsp\Linv(\xinpt-\bm{m}_k^\inp)}{\nu_k^\cond}\bm{L}_s.
	\label{eq:bgmm_unc}
\end{align}

In this work, we consider the input $\xinpt$ and the output $\xoutt$ equivalent to the state $\bm{x}$ and the control command $\bm{u}$, respectively. Note that the stability of this controller is determined by the positive-definiteness of the term $\bm{L}_k^{\out\inp}\Linv$. To guarantee the controller stability, the PoE framework is introduced in the next section. 

\subsection{Product of Experts}
\label{subsec:PoE}
Robot movements learned with state-action abstractions result in probabilistic controllers with no guarantee of stability, unless explicitly constrained to be stable as in \cite{Khansari11TRO}. To overcome this problem, we fuse the probabilistic unstable controller with another probabilistic stable controller which acts as an attractor towards the demonstration area when the uncertainty in the unstable controller is high. We refer to this fusion of controllers as a \textit{product of experts} (PoE), where each expert represents a stochastic controller with different uncertainty properties. Note that many types of controllers with different uncertainties can be fused to work in parallel. For more details, we refer the reader to~\cite{Pignat2019BayesianGM}. 

\begin{figure}[ht!]
	\centering
	\begin{subfigure}{.5\columnwidth}
		\includegraphics[width=.9\columnwidth]{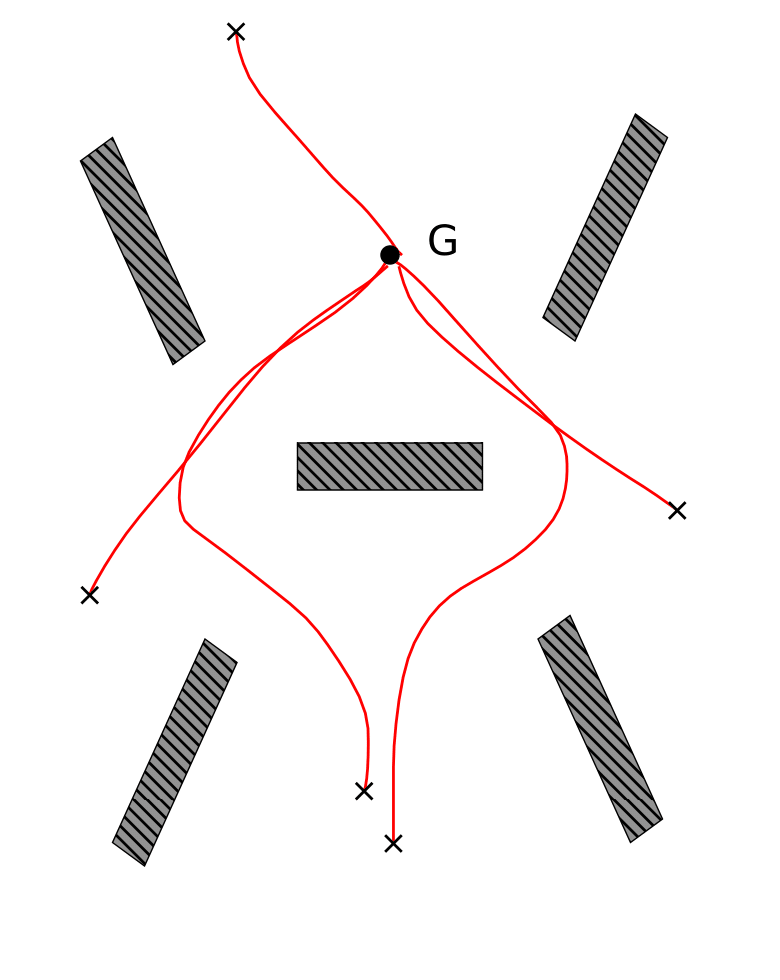}
		\caption{Initial demonstrations}
		\label{fig:initial_demos}
	\end{subfigure}%
	\hskip -2ex
	\begin{subfigure}{.5\columnwidth}
		\includegraphics[width=.9\columnwidth]{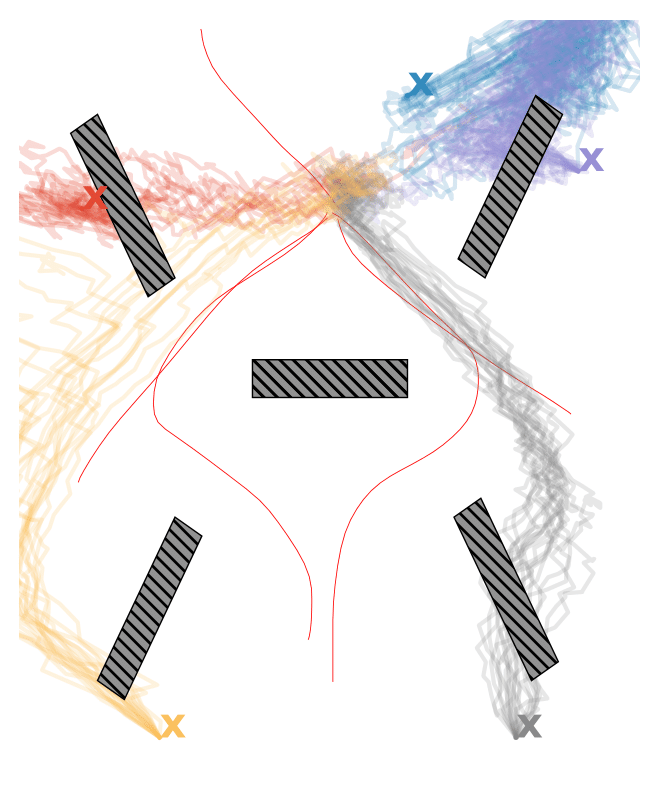}
		\caption{Policy samples from BGMM}
		\label{fig:bgmm_samples_before}
	\end{subfigure}%
	\begin{subfigure}{.5\columnwidth}
		\includegraphics[width=.9\columnwidth]{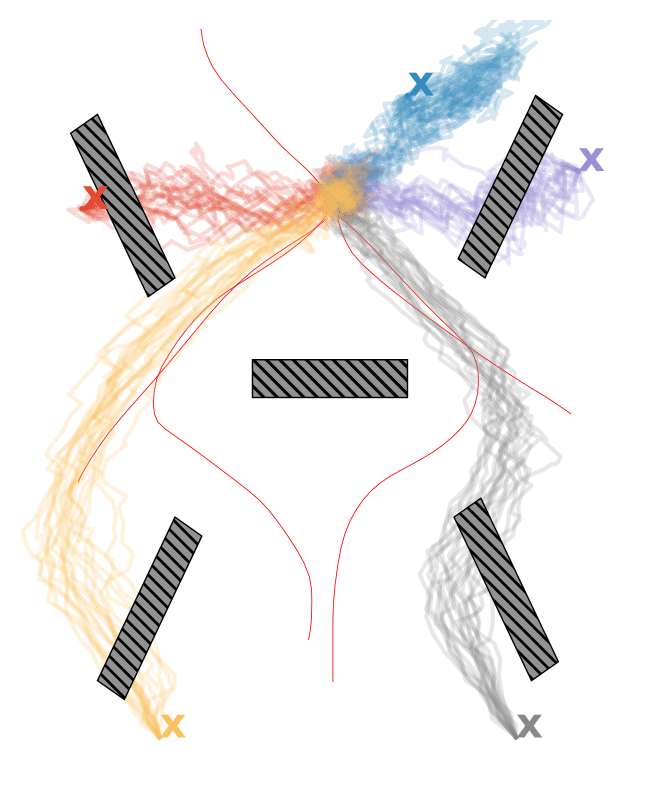}
		\caption{Policy samples from PoE}
		\label{fig:bgmm_samples_before_prod}
	\end{subfigure}
	\label{fig:initial_fig}
	\caption{\emph{(a)} Demonstrations and \emph{(b)}-\emph{(c)} reproductions of a reaching task in a cluttered environment. The goal position is denoted by G and the obstacles are represented as dashed rectangles. The demonstrated trajectories are depicted with red lines. The policy samples acquired from the BGMM and PoE are depicted by colored lines.}
\end{figure}

In this work, the stabilizing controller is defined as a probabilistic linear quadratic tracker policy, which can be expressed as a MVN. It can be viewed as a controller which attracts the system to the demonstrated regions when the BGMM controller is very uncertain. When the BGMM control policy is a GMM, the fusion or PoE is defined as the product of a GMM and a MVN, which results in another GMM policy. As an illustrative example, consider a 2D reaching task in a cluttered environment. Fig.~\ref{fig:initial_demos} displays the initial demonstrations starting from different initial positions (cross) to reach goal position ($\bm{G}$). We choose 5 different random test initial positions and reproduce the trajectories by sampling from a BGMM model and a PoE model. The resulting trajectories are shown in Fig.~\ref{fig:bgmm_samples_before} and~\ref{fig:bgmm_samples_before_prod}, respectively. Even though the trajectories are more stable in \ref{fig:bgmm_samples_before_prod} (notice that some of the trajectories in \ref{fig:bgmm_samples_before} diverge), the task cannot be accomplished without colliding with the obstacles. In this case, supplementary demonstrations are necessary, and active learning permits to collect them in an informed way.

\section{ACTIVE LEARNING WITH BAYESIAN GAUSSIAN MIXTURE MODEL}
\label{sec:ActiveLearning}
%

Control policies are defined as the probability distribution of control commands or actions $\bm{u}$ given the state $\bm{x}$, denoted as $p(\bm{u}|\bm{x})$. They encode the demonstration trajectories along with the dynamics information of the controlled system. As described in Section~\ref{subsec:BGMM}, we impose a BGMM model structure for the control policy and estimate the parameters of the predictive conditional distribution from the demonstrations. 

In this section, we present the proposed active learning of control policies approach. First, a cost function is defined using the epistemic uncertainties in the BGMM control policy and optimized while considering a soft constraint to be on the desired region of the state-space. The robot then asks for a new demonstration around the query point found by the optimization process. The data of the new demonstration is added to the previous dataset and the BGMM parameters are updated. The robot iterates this process until it reaches a predefined percentage of uncertainty reduction. 

In order to build the active learning cost function, the covariance matrices of the control policy must be decomposed into its aleatoric and epistemic parts (Section~\ref{subsec:Uncertainty}). Then, we deploy Rényi entropy to calculate epistemic uncertainties in closed-form (Section~\ref{subsec:Renyi}). The complete formulation of the resulting cost is presented in Section~\ref{subsec:Cost}. 

\subsection{Uncertainty decomposition}
\label{subsec:Uncertainty}

The uncertainty in the posterior distribution of the BGMM model encodes the variations in the demonstrations, called aleatoric uncertainty, along with the epistemic uncertainty, measuring the lack of knowledge of the model. These different uncertainty modalities are depicted in Fig.~\ref{fig:active_learning_exp} for our illustrative example. In active learning, we are interested in increasing the knowledge of the model, by providing demonstrations around interesting regions of the input space. 

In BGMM model, the covariance matrix of the conditional posterior predictive distribution of~\eqref{eq:bgmm_unc} can be decomposed into aleatoric and epistemic parts as
\begin{align}
\bm{L}_k^\cond&=\bm{L}_k^{\mathrm{al}} + \bm{L}_k^{\mathrm{ep}},
\end{align} 
where
\begin{align}
\bm{L}_k^{\mathrm{al}}&=\frac{\nu_k}{\nu_k^\cond}\bm{L}_s,\\
\bm{L}_k^{\mathrm{ep}}&=\frac{(\xinpt-\bm{m}_k^\inp)^\trsp\Linv(\xinpt-\bm{m}_k^\inp)}{\nu_k^\cond}\bm{L}_s.
\label{eq:epistemic_cov}
\end{align}
Notice that the aleatoric uncertainty does not depend on the input point $\xinpt$, while the epistemic uncertainty is a quadratic function of $\xinpt$. The former represents the variability and the noise in the demonstrations and the latter encodes the uncertainty caused by finite data. In robotics, both types of uncertainty are important to capture, i.e. the variations of the demonstrations and the uncertainty in the model, for applications such as compliance adaptation and active learning.

\subsection{Rényi entropy of the posterior distribution}
\label{subsec:Renyi}
When the posterior distribution $p(\bm{u}|\bm{x})$ is a multivariate GMM (or can be approximated by one), the information-theoretical Shannon entropy does not admit an analytical form. In order to avoid a significant amount of computational burden for the minimization of active learning cost, we use instead the quadratic Rényi entropy, which admits a differentiable closed form for GMMs~\cite{nielsen2012closed}. Another reason is that it is very close to Shannon entropy value as will be detailed below.

A random variable $\bm{U}$ from a multivariate t-distribution $\bm{U}\sim t_{\nu}(\bm{u}|\bm{\mu}(\bm{x}), \bm{\Sigma}(\bm{x}))$ can be approximated by a multivariate normal distribution with mean $\bm{\tilde{\mu}}(\bm{x})$ and covariance $\bm{\tilde{\Sigma}}(\bm{x})$ using moment-matching method, so that
\begin{alignat*}{2}
	\bm{\tilde{\mu}}(\bm{x}) = \bm{\mu}(\bm{x}), \qquad && 
	\bm{\tilde{\Sigma}}(\bm{x}) = \frac{\nu}{\nu-2}\bm{\Sigma}(\bm{x}).
\end{alignat*}
This approximation can be extended to mixtures using the same mixing coefficients. 
The Rényi entropy of order $\alpha$ is defined as $H_{\alpha}(p)=\frac{1}{1-\alpha}\text{log}\int p^{\alpha}(\bm{x})\text{d}\bm{x}$ with $\alpha>0$ and $\alpha \neq 1$. In the limit case where $\alpha \rightarrow 1$, the Rényi entropy is equivalent to the Shannon entropy defined as $H_{\alpha}(p)=-\int p(\bm{x})\text{log}p(\bm{x})\text{d}\bm{x}$. In this paper, we propose to use quadratic Rényi entropy defined as
\begin{equation*}
	H_{2}(p(\bm{u}|\bm{x}))=-\text{log}\int p^{2}(\bm{u}|\bm{x})\text{d}\bm{u},
\end{equation*}
since it admits a closed-form expression for GMMs. Note that the Rényi entropy is a non-increasing function of $\alpha$, so that $H_1(\cdot) > H_2(\cdot)$. In an active learning framework, the entropy can be used as an uncertainty measure to minimize by searching for the queries that have high entropy values. Even though the Shannon entropy is usually used in information theory, maximizing the quadratic Rényi entropy is equivalent to maximizing a lower bound of the Shannon entropy, which would also maximize it suboptimally. 
The quadratic Rényi entropy for a posterior distribution represented as a GMM $p(\bm{u}|\bm{x})= \sum_{k=1}^{K}\pi_k(\bm{x})\mathcal{N}(\bm{\mu}_k(\bm{x}), \bm{\Sigma}_k(\bm{x}))$ can be expressed as~\cite{nielsen2012closed}
\begin{equation}
	H_2(p(\bm{u}|\bm{x})) = -\log \sum_{i=1}^{K}\sum_{j=1}^{K}\pi_i(\bm{x})\pi_j(\bm{x})
	e^{\Delta_{ij}(\bm{x})},
	\label{eq:qre_gmm}
\end{equation}
where
\begin{multline}
	\Delta_{ij}
	= \frac{1}{2}
	\Big(
		\bm{\mu}_{ij}\bm{\Sigma}^{-1}_{ij}\bm{\mu}_{ij} - 
		(\bm{\mu}^\trsp_i\bm{\Sigma}^{-1}_i \bm{\mu}_i +
		 \bm{\mu}^\trsp_j\bm{\Sigma}^{-1}_j \bm{\mu}_j) \\
		- \log\frac{|\bm{\Sigma}^{-1}_i + \bm{\Sigma}^{-1}_j|}
		{|\bm{\Sigma}^{-1}_i||\bm{\Sigma}^{-1}_j|} - d\; \log2\pi
	\Big)
	\label{eq:delta}
\end{multline}
for the $i^{\text{th}}$ and $j^{\text{th}}$ components of a GMM, with $\bm{\Sigma}_{ij} = (\bm{\Sigma}^{-1}_i + \bm{\Sigma}^{-1}_j)^{-1}$ and $\bm{\mu}_{ij} = \bm{\Sigma}_{ij}(\bm{\Sigma}^{-1}_i \bm{\mu}_i + \bm{\Sigma}^{-1}_j \bm{\mu}_j)$.

\begin{figure*}[tbp]
	\centering
	\includegraphics[width=\linewidth]{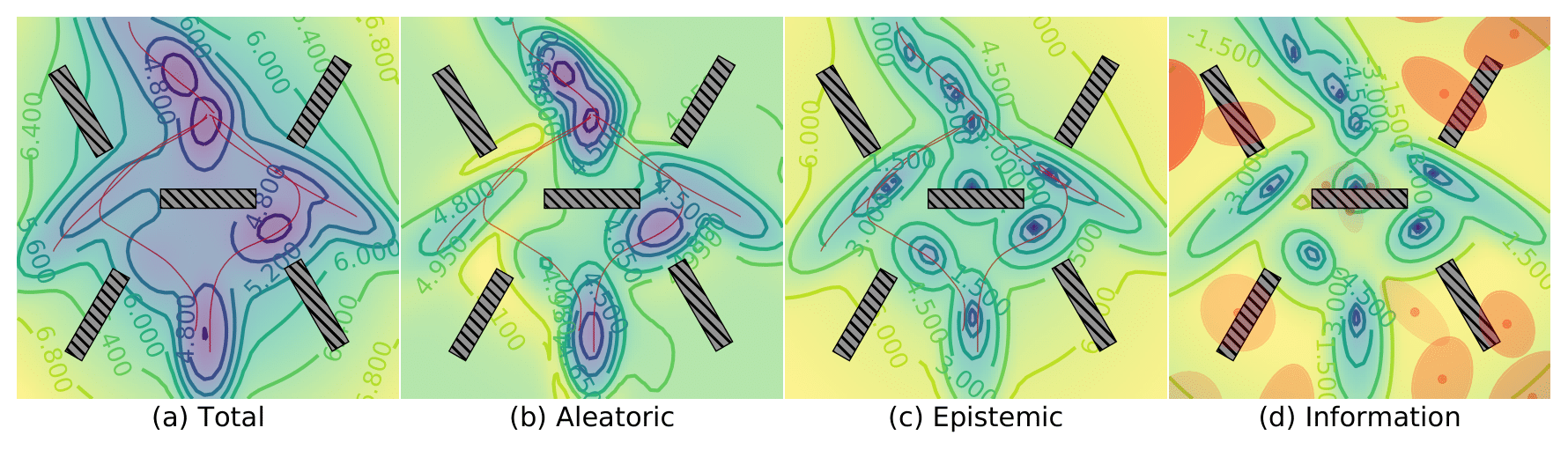}
	\vspace*{-0.7cm}
	\caption{Uncertainty colormaps of the learned control policy for a reaching task in a cluttered environment. \emph{(a)}, \emph{(b)} and \emph{(c)} show the total, aleatoric and epistemic uncertainties of the BGMM, respectively. High to low uncertainties are depicted by colors ranging from yellow to purple. \emph{(d)} depicts the information-density cost and the Gaussian components of the GMM model approximating this cost.}
	\label{fig:active_learning_exp}
\end{figure*}

Fig.~\ref{fig:active_learning_exp} depicts \emph{(a)} the total, \emph{(b)} aleatoric and \emph{(c)} epistemic uncertainties computed via the quadratic Rényi entropy of the BGMM model of our illustrative example (Fig.~\ref{fig:initial_demos}). Yellow and purple colors depict high and low uncertainties, respectively. Note that the uncertainty of the aleatoric model stays constant as we move away from known data, while it increases in epistemic model. As the epistemic model describes unseen regions, it must be used for an efficient search in the state-space.

\subsection{Information-density cost for active learning}
\label{subsec:Cost}
Following a similar approach to information weighted technique described in Section~\ref{sec:RelatedWork}, we constrain the optimization space by adding a similarity function that measures the closeness to a region of space where we want to improve our model. In this work, we represent this region as a probabilistic density function (pdf). Note that, even though the region of interest may often be represented as a uniform distribution, one may want to favor some parts of this region compared to others using other distributions. Therefore, we can solve the following optimization problem
\begin{align}
	\argmin_{\bm{x}} - H_2(p(\bm{u}|\bm{x})) - \beta\log p_{\text{sim}}(\bm{x}),
	\label{eq:id_1}
\end{align}
with the epistemic cost in closed-form, to find the next query point $\bm{x}$, where $\beta $ is a variable weighing the relative importance of the costs. In practice, uniform distributions will result in negative infinity log probabilities in the outside regions and will not have a defined gradient at the border. Therefore, we approximate the uniform distribution by an MVN using the same mean and diagonal covariance matrix to alleviate this issue. Another problem with the optimization of Eq.~\eqref{eq:id_1} is the existence of flat regions from which the optimization cannot escape. To overcome this problem, we propose to approximate the epistemic cost in Eq.~\eqref{eq:id_1} as a GMM with a variational distribution $q(\bm{x})=\sumgmm\pi_k\mathcal{N}(\xy|\bm{\mu}_k, \bm{\Lambda}_k^{-1})$ to represent all the regions where epistemic uncertainty is high, using reverse KL divergence as in
\begin{align}
	\argmin_{\bm{x}} KL\Big(q(\bm{x})||H_2(p(\bm{u}|\bm{x}))+\beta\log p_{\text{sim}}(\bm{x})\Big).
	\label{eq:id_2}
\end{align}
Note that one can also augment the epistemic cost defined in Eq. \eqref{eq:id_1} with other costs (see robotic experiment in Section V.B.), so that $q(\bm{x})$ can represent a more constrained space (e.g. being away from an undesirable region). We can obtain the next query point either by sampling from $q(\bm{x})$ or by taking the mean of one of the components. As we add more demonstrations and improve our model using this query point, the optimization in Eq.~\eqref{eq:id_2} can be initialized with the parameters of the previous $q(\bm{x})$, which would increase convergence speed. We expect a decrease of entropy in $q(\bm{x})$ at each iteration of active learning. This gives us a natural way of monitoring the uncertainty reduction. 

Fig.~\ref{fig:active_learning_exp}\emph{d} shows the information density colormap favoring to be inside of the figure frame where we want to generalize our model. It also shows the GMM contour ellipses (with 1 standard deviation) which approximate the high information-density regions (yellow). The transparency reflects the mixing coefficient of the GMM. We can observe that the highly uncertain regions are well approximated.

\section{EXPERIMENTS}
\label{sec:Experiments}
\subsection{Illustrative reaching task}
We use the proposed active learning framework to gather iteratively 10 more demonstrations for our illustrative 2D reaching task. At each step, the model informs the teacher on the next query point, given by the mean of the GMM component with the highest mixing coefficient (corresponding to the highest uncertainty). As any sample from that component can be used as a next query point, the closest feasible position to the mean can be chosen if the mean does not correspond to a feasible location, e.g. if it collides with the obstacles. 

\begin{figure}[tbp]
	\centering
	\includegraphics[width=1.\columnwidth]{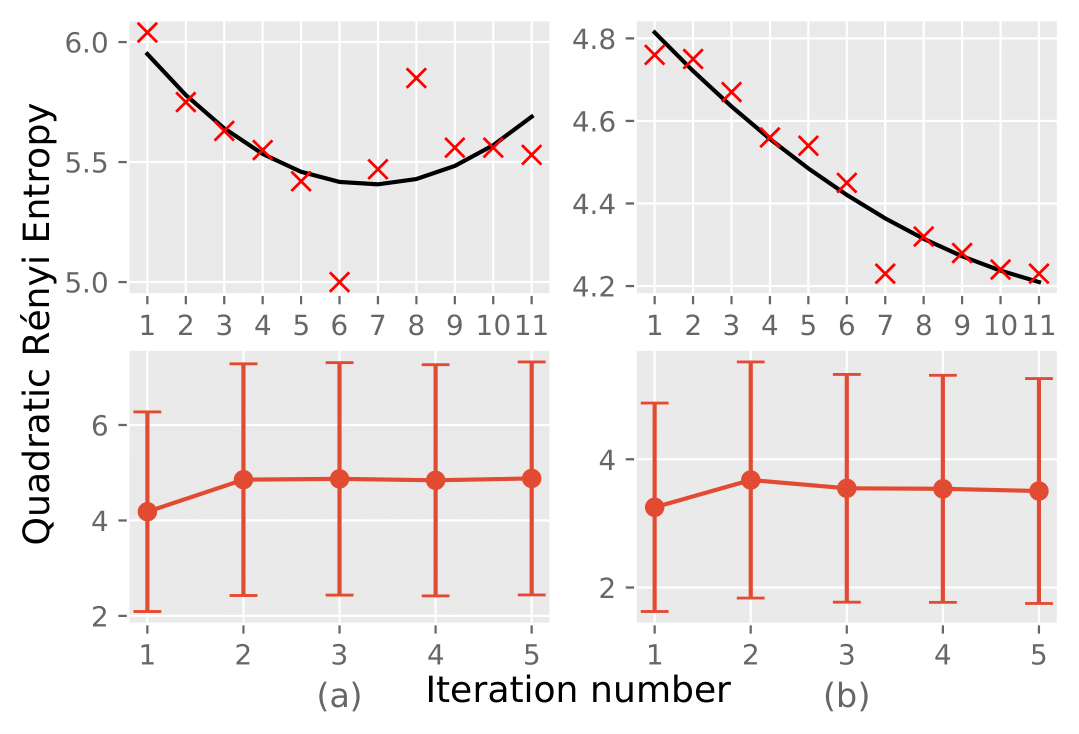}
	\caption{  Evolution of the quadratic Rényi entropy of (a) the GMM model that approximates highly uncertain regions and (b) the marginal BGMM model. Top figures represent the evolution for the proposed active learning, while the error bars in bottom figures show the mean and the standard deviation of 5 different random exploration for 5 iterations. }
	\label{fig:all_entropy}
\end{figure}

Note that we are interested in reducing the epistemic uncertainties in the conditional model, which is a function of the input point as in \eqref{eq:epistemic_cov}. In order to define an entropy reduction, we need a measure that does not depend on the input point. We can thus measure how much the entropy changes via the GMM model which approximates highly uncertain regions. Fig.~\ref{fig:all_entropy}\emph{a}-\emph{(top)} shows the evolution of the quadratic Rényi entropy of the GMM model across the active learning iterations. Red crosses show the current entropy values, whereas the black curve is 2D polynomial fit to these values. We can observe that the entropy of the GMM is reduced until there is no component left which can specialize in certain regions with small covariance (small covariance means low entropy). After $6$ iterations, the entropy starts to increase as the components are more diffused with bigger covariance matrices. We generally observed that the entropy of the GMM behaves similarly to the black curve in Fig.~\ref{fig:all_entropy}\emph{a}-\emph{(top)}.
The evolution of the entropy of the marginal model $p(\bm{x})$ is represented in Fig.~\ref{fig:all_entropy}\emph{b}-\emph{(top)}. As expected, the entropy of the marginal model decreases with the quantity of data. Therefore, it results in no explicit method to infer the convergence of the learning process. In contrast, with our GMM model, one can argue that the system has learned a significant percentage of the unseen regions after $6$ iterations. 

We conducted 5 more experiments performing active learning where new random demonstrations are provided for 5 iterations. The mean and standard deviation of 5 experiments at each iteration are shown in Fig.~\ref{fig:all_entropy}\emph{a}-\emph{(bottom)} for the GMM model and in Fig.~\ref{fig:all_entropy}\emph{b}-\emph{(bottom)} for the marginal model. This demonstrates that the random exploration is not guaranteed to reduce the epistemic uncertainties, even in the marginal model.

\begin{figure}
\centering
\begin{subfigure}{.49\columnwidth}
	\includegraphics[width=.9\columnwidth]{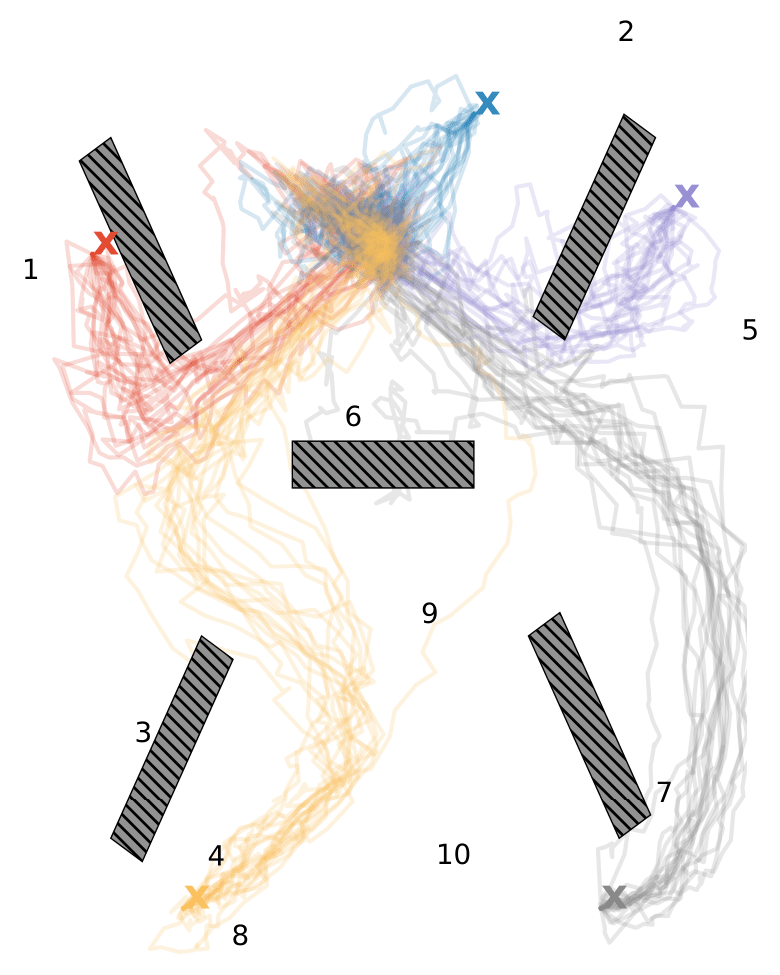}
	\caption{Policy samples from BGMM}
	\label{fig:bgmm_samples_after}
\end{subfigure}%
\hskip -1ex
\begin{subfigure}{.49\columnwidth}
	\includegraphics[width=.9\columnwidth]{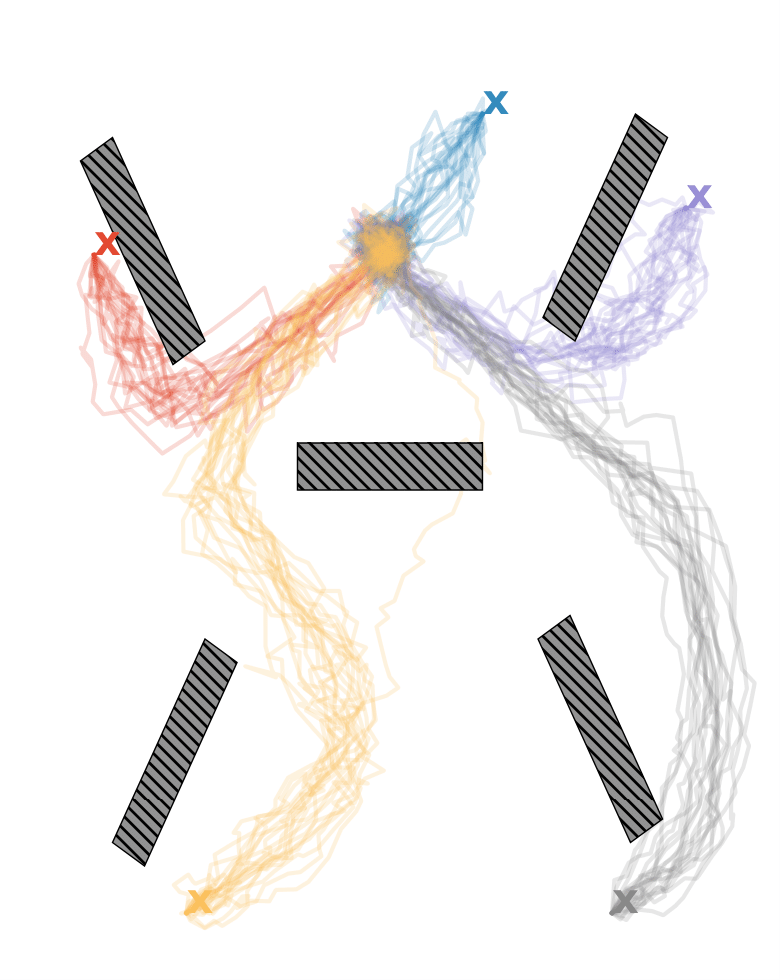}
	\caption{Policy samples from PoE}
	\label{fig:bgmm_samples_after_poe}
\end{subfigure}
\caption{Reproductions of the learned policy after 10 iterations of active learning. The numbers on (a) denotes the location of the query point at each iteration of active learning.}
\label{fig:repro_after}
\end{figure}

The resulting reproductions from the chosen random initial test positions using samples from the updated BGMM and PoE policies are shown in Fig.~\ref{fig:bgmm_samples_after} and Fig.~\ref{fig:bgmm_samples_after_poe}, respectively. We observe that both policies successfully avoid all the obstacles in the average, while using PoE framework results in a more stable system. The query points of each iteration of active learning are also labelled in Fig.~\ref{fig:bgmm_samples_after}. We observe that these query points are rather intuitive, as they correspond to locations that could be chosen by a human to better teach the task to the robot. In contrast, informative query points may be very difficult to choose in other cases where the query space is not easily interpretable.

\subsection{Robot Experiment}
We investigate the reaching in a cluttered environment task shown in Figure \ref{fig:setup} within our active learning framework. The main challenge of this task is to place the cup inside the white box without colliding with the environment and without pouring the cup. The robot can place the cup from any open side of the box, as long as the cup is inside. Planning methods can be applied to find a joint configuration trajectory starting from a given initial configuration of the robot without colliding with the environment, given the size and positions of the obstacles. However, learning control policies using BGMM offers the advantage of sampling the next state much faster than standard planning methods. It also provides a formal way of improving the planned trajectory using active learning framework proposed in this paper. For the improvement of the learned policy, it is difficult for the teacher to choose informative joint configurations intuitively as the demonstrations can take place starting from many different end-effector positions, which correspond to many more joint configurations. Our goal in this experiment is to show that our method provides \textquotedblleft intuitive\textquotedblright~ and informative query points in the joint space of the robot. 

\begin{figure}
	\includegraphics[width=\columnwidth]{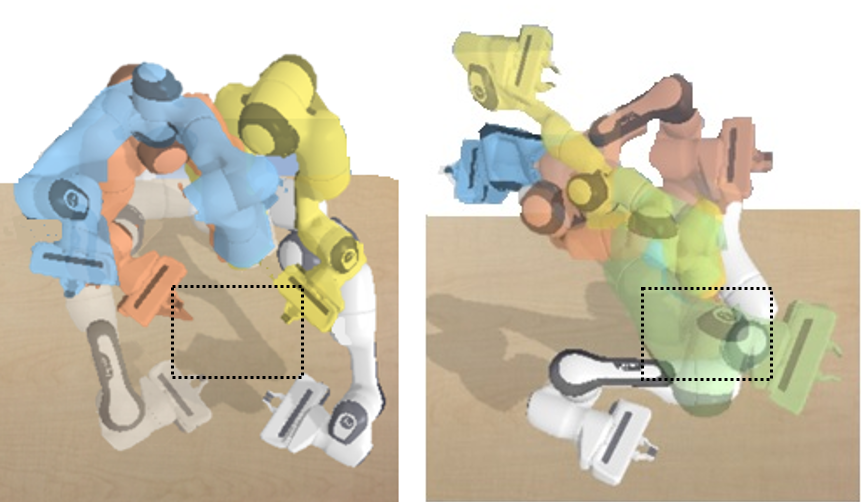}
	\caption{ (Left) Initial configurations of the demonstrations, (Right) Requested initial configurations for demonstration}
	\label{fig:robots}
\end{figure}

We first demonstrate the reaching task from 11 different initial configurations and learn our control policy. Note that the demonstrations are taken from each side of the box, where it was easy to perform kinesthetic teaching. The initial configurations of the demonstrations are depicted in Figure \ref{fig:robots} (left).

\vspace{-0.2cm}

To improve the model, one need to start from a rather different and informative initial configuration of the arm, which is not easy. Note that the robot has to maintain upright position of the cup to place it inside the box without pouring it and without colliding with the environment. That is why the search space we are interested in is constrained such that we add 2 more cost functions to Eq. \eqref{eq:id_1} in the form of probability distributions: \textit{i}) a cost to keep orientation with respect to x-y axis of the robot base fixed   and  \textit{ii}) a cost to be within the joint limit range of the robot as

\begin{align*}
p(\bm{x}) &= H_2(p(\bm{u}|\bm{x}))+\beta \text{f}_{\text{limits}}(\bm{x}) + \alpha\log p_{\text{upright}}(\bm{x})
\end{align*}
where $\text{f}_{\text{limits}}(\cdot)$ is typically the sum of lognormal cumulative distribution functions representing inequality functions which represent the joint limits and $p_{\text{upright}}(\cdot)$ is a normal distribution on the quaternion describing upright orientation in the manifold.

We approximate this cost by a GMM of 10 components minimizing KL divergence between $q(\bm{x})$ and $q(\bm{x})$ as in Eq. \ref{eq:id_2}. The resulting query configurations (samples from GMM) are given in Figure \ref{fig:robots} (right). We can see that our GMM could in fact approximate highly uncertain and unseen configurations of the robot as it requests demonstrations around these regions. These  configurations are also within the joints range of the robot, and maintain approximately a fixed x-y axis orientation so that the robot will keep the cup upright, without pouring. Although showing that the usefulness of encoding aleatoric uncertainties here is out of scope of this paper, it has been exploited in the previous work in \cite{Pignat2019BayesianGM}. Since the aperture size of the sides are big enough, one can imagine exploiting high variations in the demonstrations while the end-effector enters one side of the box. The learned model would create compliant control commands in these areas which would help the teacher to correct the robot movement during a failing execution. Note that GPR could not encode aleatoric uncertainties.

\section{CONCLUSION}
This paper presented a novel active learning framework allowing a robot to ask for informative new demonstrations. The presented framework is based on an information-density cost built from a representation of the epistemic uncertainties of a BGMM model. A closed-form cost solution for GMMs can be obtained thanks to the properties of the quadratic Rényi entropy. New query points can then be efficiently obtained by maximizing a GMM approximation of the proposed active learning cost. Our experiments showcase that our approach allows a robot to improve its representation of a task, as well as its corresponding generalization capabilities. 

The model in our work can assess the uncertainty of the control command given the current state. However, in many application in robotics, it is necessary to propagate these uncertainties to determine the uncertainty on the whole trajectory. Future work should focus on either how to propagate uncertainties in the state-action policies, or on extending it to trajectory policies. Another future work consists in extending our results to theoretically determine a threshold to stop the learning process, which in turn would be useful for determining a sufficient number of demonstrations so that the model can generalize the fastest in the desired space. We believe that the framework can then be used to answer two of the main questions of LfD, which are \textit{i}) Where to give demonstrations? and \textit{i}) How many demonstrations are required?.




\bibliographystyle{IEEEtran}
\bibliography{arxiv_iros2020_girgin}

\end{document}